\pdfoutput=1
\documentclass[11pt,a4paper]{article}
\usepackage[hyperref]{acl2018}
\usepackage{times}
\usepackage{latexsym}

\usepackage{amsmath}
\usepackage{graphics}
\usepackage{tabularx}

\usepackage{url}
\usepackage{booktabs}
\usepackage[normalem]{ulem}

\makeatletter
\DeclareRobustCommand{\textsupsub}[2]{{%
		\m@th\ensuremath{%
			^{\mbox{\fontsize\sf@size\z@#2}}%
			_{\mbox{\fontsize\sf@size\z@#1}}%
		}%
}}

\DeclareRobustCommand{\textsupsubMaxBf}[2]{{%
		\m@th\ensuremath{%
			^{\mbox{\textbf{\fontsize\sf@size\z@#2}}}%
			_{\mbox{\fontsize\sf@size\z@#1}}%
		}%
}}

\makeatother

\newcommand{\resultSt}[2]{#1$\pm$#2}

\definecolor{BoxCol}{cmyk}{.3,.05,0,0} 
\definecolor{unimittelblau}{cmyk}{1,.4,0,0} 
\definecolor{unimittelblauhell}{cmyk}{.3,.05,0,0} 
\definecolor{unianthrazithell}{cmyk}{.16,.06,.06,.26} 
\definecolor{unianthrazitheller}{cmyk}{.16,.06,.06,.22}
\definecolor{unimittelblau100}{cmyk}{1,.4,0,0}
\definecolor{unimittelblau70}{cmyk}{.7,.2,0,0}
\definecolor{unimittelblau50}{cmyk}{.5,.1,0,0}
\definecolor{unimittelblau30}{cmyk}{.3,.05,0,0}
\definecolor{unianthrazit100}{cmyk}{.5,.2,.2,.85}
\definecolor{unianthrazit70}{cmyk}{.35,.14,.14,.6}
\definecolor{unianthrazit50}{cmyk}{.25,.1,.1,.43}
\definecolor{unianthrazit30}{cmyk}{.15,.06,.06,.26}
\definecolor{univiolettblau100}{cmyk}{.9,.6,0,0}
\definecolor{univiolettblau70}{cmyk}{.68,.4,0,0}
\definecolor{univiolettblau50}{cmyk}{.45,.26,0,0}
\definecolor{univiolettblau30}{cmyk}{.27,.12,0,0}
\definecolor{unigelb}{cmyk}{0,.1,1,0} 
\definecolor{uniorange}{cmyk}{0,.7,1,0}
\definecolor{unirot}{cmyk}{0,1,1,0} 
\definecolor{unipink}{cmyk}{0,1,0,0}
\definecolor{univiolett}{cmyk}{.6,1,0,0}
\definecolor{unidunkelviolett}{cmyk}{0.8,0.8,0,0}
\definecolor{unimittelviolett}{cmyk}{0.8,1,0,0}
\definecolor{unituerkis}{cmyk}{1,0,.4,0} 
\definecolor{uniapfelgruen}{cmyk}{.5,0,1,0} 
\definecolor{unianthrazitmittel}{cmyk}{.25,.1,.1,.43}    

\definecolor{middlegreen}{cmyk}{.6,0,1,0}
\definecolor{darkgreen}{cmyk}{0.9,0,1,0.60}
\definecolor{darkpink}{cmyk}{0,1,0.31,0.34}

\colorlet{citecolor}{unituerkis}
\definecolor{infcolor}{cmyk}{0,.7,1,0}
\colorlet{univiolettMiddle}{univiolett!80!white}
\colorlet{infcolorLight}{infcolor!50!white}
\colorlet{univiolettLight}{univiolett!40!white}
\colorlet{lightgray}{unianthrazitheller!20!white}

\colorlet{phdBoxColor}{unimittelblauhell!40}
\colorlet{avgPoolingColor}{univiolettLight}
\colorlet{wghPoolingColor}{unimittelblau!50}
\colorlet{baselineColor}{unianthrazithell!70}
\colorlet{bestColor}{unirot}
\colorlet{Aggregatecolor}{unimittelviolett!55!white}
\colorlet{CNNcolor}{univiolett}
\colorlet{RNNcolor}{unidunkelviolett}
\colorlet{wordcolor}{univiolettblau100} 
\colorlet{charcolor}{uniorange}
\colorlet{okcolor}{uniapfelgruen!55!white}
\colorlet{template1color}{univiolettblau100}
\colorlet{template2color}{darkpink}

\aclfinalcopy 

\title{Sequence-to-Sequence Models for Data-to-Text Natural Language Generation: Word- vs. Character-based Processing and Output Diversity}

\author{First Author \\
	Affiliation / Address line 1 \\
	Affiliation / Address line 2 \\
	Affiliation / Address line 3 \\
	{\tt email@domain} \\\And
	Second Author \\
	Affiliation / Address line 1 \\
	Affiliation / Address line 2 \\
	Affiliation / Address line 3 \\
	{\tt email@domain} \\}

\author{
  Glorianna Jagfeld, Sabrina Jenne, Ngoc Thang Vu\\
  Institute for Natural Language Processing (IMS) \\
  Universit\"at Stuttgart, Germany \\
  {\tt \{jagfelga,beersa,thangvu\}@ims.uni-stuttgart.de}}

\date{}

\begin{document}
	\maketitle
	\begin{abstract}
		We present a comparison of word-based and character-based sequence-to-sequence models for data-to-text natural language generation, which generate natural language descriptions for structured inputs.
		On the datasets of two recent generation challenges, our models achieve comparable or better automatic evaluation results than the best challenge submissions.
        Subsequent detailed statistical and human analyses shed light on the differences between the two input representations and the diversity of the generated texts.
        In a controlled experiment with synthetic training data generated from templates, we demonstrate the ability of neural models to learn novel combinations of the templates and thereby generalize beyond the linguistic structures they were trained on.
	\end{abstract}

\section{Introduction}

Natural language generation (NLG) is an actively researched task, which according to~\citet{NLGSurvey_Gatt18} can be divided into text-to-text generation, such as machine translation~\citep{NMTChapter_Koehn17}, text summarization~\citep{GetToThePointSummarization_See17}, or open-domain conversation response generation~\citep{NeuralConvModel_Vinyals15} on the one hand, and data-to-text generation on the other hand.
Here, we focus on the latter, the task of generating textual descriptions for structured data.
Data-to-text generation comprises the generation of system responses based on dialog acts in task-oriented dialog systems~\citep{SemanticLSTMNLG_Wen15}, sport games reports and weather forecasts~\citep{RobocubWeatherGov_Angeli10}, and database entry descriptions~\citep{WebNLGTrainingCorporaNLGMicroPlanners_Gardent17}.
In this paper, we focus on sentence planning and surface realization. We build on data-to-text datasets of two recent shared tasks for end-to-end NLG, namely the E2E challenge~\citep{E2E_Novikova_2017} and WebNLG challenge~\citep{WebNLGReport_Gardent17}.
Example input-text pairs for both datasets are shown in Figure~\ref{fig:examples}.

\begin{figure*}[tb]
    \textbf{E2E input:} name[Midsummer House], customer rating [average], near [The Bakers]\\
    \textbf{reference 1:} Customers gave Midsummer House, near The Bakers, a 3 out of 5 rating.\\
    \textbf{reference 2:} Midsummer house has an average customer rating and is near The Bakers.\\
    \textbf{delexicalized input:} name[\textsc{name}], customer rating [average], near [\textsc{near}]\\
    \textbf{delexicalized reference 1:} Customers gave \textsc{name}, near \textsc{near}, a 3 out of 5 rating.
    \vspace{2mm}
    
	\textbf{WebNLG input:}
	cityServed(Abilene Regional Airport[Abilene]), isPartOf(Abilene[Texas])\\
	\textbf{reference 1:} Abilene is in Texas and is served by the Abilene regional airport.\\
	\textbf{reference 2:} Abilene, part of Texas, is served by the Abilene regional airport.\\
    \textbf{delexicalized input:} city served(\textsc{agent-1[bridge-1]}), is part of(\textsc{bridge-1[patient-1]})\\
    \textbf{delexicalized reference 1:} \textsc{bridge-1} is in \textsc{patient-1} and is served by the \textsc{agent-1}.

	\caption{Example input-reference pairs from the E2E and WebNLG development set.}
    \label{fig:examples}
\end{figure*}

Neural sequence to sequence (Seq2Seq) models~\citep{Seq2Seq_Graves13,Seq2Se2_Sutskever14} have shown promising results for this task, especially in combination with an attention mechanism~\citep{Attention_Bahdanau14,Attention_Luong15}.
Several recent NLG approaches~\citep{Seq2Tree_Dusek16,SelectiveNLG_Mei_16,GloballyCoherent_NLG_Kiddon16,Char2Char4E2E_Shubham_2017}, as well as most systems in the E2E and WebNLG challenge are based on this architecture.
While most NLG models generate text word by word, promising results were also obtained by encoding the input and generating the output text character-by-character~\cite{CharacterLevelGenerative_Lipton15,CharNLG_Finite_State_Goyal_16,Char2Char4E2E_Shubham_2017}.
Five out of 62~E2E challenge submissions operate on the character-level.
However, it is difficult to draw conclusions from the challenge results with respect to this difference, since the submitted systems also differ in other aspects and were evaluated on a single dataset only.

Besides adequacy and fluency, variation is an important aspect in NLG~\cite{NLGEvalVariation_Stent05}.
In addition to comparing the linguistic and contentwise correctness of word- and character-based Seq2Seq models through automatic and human evaluation, we investigate the variety of their outputs.
While template-based systems can assure perfect content and linguistic quality, they often suffer from low diversity.
Conversely, neural models might generalize beyond a limited amount of training texts or templates, thereby producing more diverse outputs.
To test this hypothesis, we train Seq2Seq models on template-generated texts with a controlled amount of variation and show that they not only reproduce the templates, but also generate novel structures resulting from template combinations.

In sum, we make the following contribution:
\begin{itemize}
	\item We compare word- and character-based Seq2Seq models for NLG on two datasets.
	\item We conduct an extensive automatic and manual analysis of the generated texts and compare them to human performance.
    \item In an experiment with synthetic training data generated from templates, we demonstrate the ability of neural NLG models to learn template combinations and thereby generalize beyond the linguistic structures they were trained on.
\end{itemize}


\section{Related Work}
This section reviews relevant related work according to the two main aspects of this paper: different input and output representations for data-to-text NLG as well as measuring and controlling the variation in the generated outputs.

\subsection{Input and Output Representations}
While the first NLG systems relied on hand-written rules or templates that were filled with the input information~\citep{NLGTemplateDialog_Cheyer06,NLGTemplateDialog_Mirkovic06}, the availability of larger datasets has accelerated the progress in statistical methods to train NLG systems from data-text pairs in the last twenty years~\citep{NLGCorpusBasedOvergeneration_Oh00,NLGCorpusBasedFactoredLM_Mairesse14}.
Generating output via language models based on recurrent neural networks~(RNNs) conditioned on the input~\citep{GeneratingTextsRNN_Sutskever11} proved to be an effective method for end-to-end NLG~\citep{StochasticLanguageGenerationRNNCNNReranking_Wen15,SemanticLSTMNLG_Wen15,MultiDomainSC_LSTM_Wen_16}.

The input can be represented in several ways: (1) In a discrete vector space via one-hot-vectors~\citep{StochasticLanguageGenerationRNNCNNReranking_Wen15,SemanticLSTMNLG_Wen15}, or in a continuous space either (2) by encoding fixed-size input information in a feed-forward neural network~\citep{LearningToGenerateProductReviews_Zhou17,ChallengesInData2DocGeneration_Wiseman17} or (3) by the means of an encoder RNN, which processes variable-sized inputs sequentially, giving rise to the Seq2Seq architecture.

Character-based Seq2Seq models were first proposed for neural machine translation~\cite{CharBasedNMT_Ling15,Bpe2Char_Chung16,Char2CharNMT_Lee17}.
Their main advantage over word-based models is that they can represent an unlimited word inventory with a small vocabulary.
They can learn to copy any string from the input to the output, which is especially useful for data-to-text NLG, as information from the input such as the name of a restaurant or a database entity is often expected to appear verbatim in the generated text.
Word-based models, in contrast, have to make use of delexicalization during pre- and postprocessing~\citep{SemanticLSTMNLG_Wen15,Seq2Tree_Dusek16} or have to apply dedicated copy mechanisms~\citep{CopyingInSequenceToSequence_IncorporatingGu16,GetToThePointSummarization_See17,ChallengesInData2DocGeneration_Wiseman17} to handle open vocabularies.
The other side of the coin is that sequences are much longer in character-based processing, implying longer dependencies and more computation steps for encoding and decoding.

Subword-based representations~\citep{SubwordUnits_Sennrich16,NMTGoogle_Wu16} can offer a trade-off between word- and character-based processing and are a popular choice in NMT and summarization~\citep{GetToThePointSummarization_See17}.
Here, the vocabulary consists of subword units of different lengths, which are assigned by minimizing the entropy on the training set.
We also experimented with such representations in preliminary experiments, but found them to perform much worse than word- or character-based representations.
Our impression is that recurring entity names in the training data coming from multiple reference texts for the same input lead to overfitting on the training vocabulary and to poor generalization to novel inputs.
This is also reflected by the rather unsatisfying performance of subword-based approaches in the E2E\footnote{The subword-based \textit{bzhang\_submit} system has the second best ROUGE-L score, but ranks poorly in terms of BLEU and quality in the human evaluation, see~\url{http://www.macs.hw.ac.uk/InteractionLab/E2E/\#results}.} and WebNLG challenge (ADAPT system~\cite{WebNLGReport_Gardent17}).

\subsection{Output Diversity}
Evaluation of data-to-text NLG has traditionally centered around semantic fidelity, grammaticality, and naturalness~\citep{NLGSurvey_Gatt18,PersonalityVariationNeuralNLG_Oraby18}. More recently, the controllability of the style of the outputs and their variation has moved into focus as well~\citep{ControlingLinguisticStyleNLG_Ficler_17,ResponseGenerationCustomerSerivePersonality_Herzig17,PersonalityVariationNeuralNLG_Oraby18,MultiVoice4PersonalityNeuralDialog_Oraby18}.

\citet{PersonalityVariationNeuralNLG_Oraby18} showed that the n-gram entropy of the outputs of a neural NLG system is significantly lower compared to its training data.
This can be seen as evidence that the NLG system extracts only a few dominant patterns from the training data that it will generate over and over.
Without explicit supervision signals, neural NLG models cannot distinguish linguistic or stylistic variation from noise.
In the context of image caption generation \citet{LM4ImageCaptioning_Devlin15} found Seq2Seq models to exactly reproduce sentences from their training data for 60\% of the test instances.

Several approaches have been proposed to control NLG outputs with respect to certain stylistic aspects, e.g., mimicking a specific persona or character~\citep{FilmCharStage_Lin11,FilmCharacterNLG_Walker11,Persona_Li16}, personality traits~\citep{Big5PersonalityTraits_Mariesse08,ResponseGenerationCustomerSerivePersonality_Herzig17,PersonalityVariationNeuralNLG_Oraby18,MultiVoice4PersonalityNeuralDialog_Oraby18}, or various linguistic aspects such as formality, voice, descriptiveness~\citep{ControlingLinguisticStyleNLG_Ficler_17,StyleMT_Bawden17,FormalStyleMT_Niu17}.
All share the feature that the NLG model is conditioned on a representation of the desired aspect in addition to the usual semantic input representation.
While this approach makes it possible to successfully control particular, clearly defined aspects of the generated texts, further research is needed to grant more flexible and comprehensive NLG output control.

\section{Models}
To encode variable-length inputs and generate variable-length texts, we implement a standard Seq2Seq model~\citep{Seq2Seq_Cho14} with Long Short-Term Memory (LSTM) cells~\cite{HochreiterLSTM_97} and attention.
Given a training dataset of input-text pairs $D = \{(x^1, \bar{y^1}), (x^2, \bar{y^2}) \dots \}$, the model encodes an input sequence $x = \{x_1 \dots x_n\}$ of symbols $x_i$ into a sequence of hidden states $\{h_1 \dots h_n\}$ by applying a recurrent neural network (RNN) with LSTM cells that can store and forget sequence information:

\begin{equation}
	h_t =\text{LSTM}(X^{\text{in}}x_t, h_{t-1})
\end{equation}

The decoder generates the output sequence~$y_1 \dots y_m$ one symbol~$y_t$ at a time by computing $p(y_t|y_1 \dots y_{t-1}, x) = \text{softmax}(W^{\text{out}}(c_t))$.

The decoder output $c_t$, also referred to as context vector, summarizes the input information in each decoding step as weighted sum of the encoder hidden states: $c_t = \sum_{i=1}^n \alpha_{ti}h_i$. The attention weights~$\alpha_{ti}$ are computed with the general attention mechanism $\alpha_{ti} = \text{softmax}(s_t W^\text{a} h_i)$~\citep{Attention_Luong15}.
The decoder hidden states $s_t$ are computed recursively based on the previous output token and decoder output:

\begin{equation}
	s_t = \text{LSTM}((X^{\text{out}}(y_{t-1}) \circ c_{t-1}), s_{t-1})
\end{equation}

$s_0$ is initialized to the final encoder hidden state $h_n$, $h_0, c_{-1}$ are initialized to~0; $\circ$ denotes concatenation.
The parameters of the models are the input and output embedding matrices $X^{\text{in}}$, $X^{\text{out}}$, the encoder and decoder LSTM parameters, the attention matrix $W^\text{a}$ and the output matrix $W^{\text{out}}$.
They are optimized by minimizing the cross entropy of the generated texts $y^j$ with the given references $\bar{y^j}$ for each example in the training set.

Instead of forcing the decoder to decide on a single output symbol in each decoding step, we apply beam search~\cite{Seq2Seq_Cho14,Attention_Bahdanau14} to explore $n$-best partial hypotheses in parallel.

In the word-based model, each input symbol~$x_t$ and output symbol~$y_t$ denotes a token.
In contrast, in the character-based model, each input and output symbol denotes a single character.
Our models learn separate encoder and decoder embedding matrices.

\section{Data}
We use two recently collected crowd-sourced data-to-text datasets since they are larger and offer more linguistic variety than previously available datasets~\citep{E2E_Novikova_2017,WebNLGTrainingCorporaNLGMicroPlanners_Gardent17}.
The E2E dataset~\citep{E2E_Novikova_2017} consists of 47K~restaurant descriptions based on 5.7K~distinct inputs of 3-8~attributes (\textit{name, area, near, eat type, food, price range, family friendly, rating}), split into 4862~inputs for training, 547~for development and 630~for testing.
The WebNLG dataset~\citep{WebNLGTrainingCorporaNLGMicroPlanners_Gardent17} contains 25K~verbalizations of 9.6K~inputs composed of 1-7~DBpedia triples from 15~categories such as \textit{athletes, comic characters, food, sport teams}. It is divided into 6893~inputs for training, 872~for development and 1862~for testing.
Both datasets have multiple verbalizations for each input.
On average there are 8.3~(min.~1, max.~46) verbalizations per input in the E2E dataset and 2.63 (min.~1, max.~12) in the WebNLG dataset, respectively.

To preprocess both datasets, we lowercase all inputs and references and represent the inputs in the bracketed format as shown in  Figure~\ref{fig:examples}.
For the word-based processing we additionally tokenize the texts with the nltk-tokenizer~\citep{NLTK_Bird09:NLP} and apply delexicalization, as also illustrated in Figure~\ref{fig:examples}.
For the E2E dataset we adopt the challenge's baseline delexicalization strategy~\citep{Seq2Tree_Dusek16}, which replaces the values of the two open-class attributes \textit{name} and \textit{near} in the input and references by placeholders.
For the WebNLG dataset, we adopt the delexicalization strategy of the \textsc{Tilburg} submissions to the challenge, since it performed well and does not require external information.
They replaced the subject and object entities of the DBpedia triples in the input and text by numbered placeholders \textsc{agent-n, patient-n, bridge-n}, depending on whether they only appear as subject, object or in both roles in the input of an instance.
Additionally, we split properties at the camel case in this dataset for both the word- and character-based models as proposed by the \textsc{Adapt} and \textsc{Melbourne} submissions.
Table~\ref{tab:dataset} displays statistics for both datasets and processing types.

\begin{table}
	\resizebox{\linewidth}{!}{
\begin{tabular}{lrrrr}
	& \multicolumn{2}{c}{E2E} & \multicolumn{2}{c}{WebNLG} \\
	& word & char. & word & char. \\ \midrule
	avg. input length	& 28.5 & 106.0 & 24.8 & 139.8\\
	avg. text length	& 20.0 & 109.3 & 18.8 & 117.1\\
	input vocabulary	& 48 & 39 &  312& 78\\
	output vocabulary	& 2,721 & 53 & 4,264& 83\\
\end{tabular}
}
	\caption{E2E and WebNLG training split statistics for word-based processing after delexicalization and character-based processing.}
   	\label{tab:dataset}
\end{table}

\section{Experiments}
We conduct our experiments with the OpenNMT toolkit~\citep{OpenNMT_Klein17}, which we extend to also perform character-based processing.
We tuned the hyperparameters for each dataset and processing method to optimize the BLEU score on the development sets.
The word-based model for the E2E~dataset is trained by stochastic gradient descent (SGD)~\cite{SGD_Robbins51} and an initial learning rate of~1.0.
For all other models, we achieved better performance with the Adam optimizer~\cite{Adam_Kingma15} with an initial learning rate of 0.001.
If there is no improvement in the development perplexity, or in any case after the eighth~epoch, we halve the learning rate.
Also, we clip all gradients to a maximum of five.
We use a batch size of~64.
To prevent overfitting, we drop out units in the context vectors with a probability of~0.3.
We keep the model with the lowest development perplexity in 13~training epochs.

The word-based E2E model has 64-dimensional word embeddings and a single encoder and decoder layer with 64~units each.
All other models use 500-dimensional word- or character embeddings and two layers in the encoder and decoder with 500~dimensions each.
While a unidirectional encoder was sufficient for the word-based models, bidirectional encoders were beneficial for the character-based models on both datasets.

We use a beam size of~15~for decoding with the word-based models, and found a smaller beam of five to yield better results for the character-based models.
This is probably due to the much smaller vocabulary size of the character-based models.

For automatic evaluation, we report BLEU~\citep{BLEU_Papineni02}, which measures the precision of the generated n-grams compared to the references, and recall-oriented ROUGE-L~\citep{ROUGE-L_Lin04}, which measures the longest common subsequence between the generated texts and the references.
We compute these scores with the E2E challenge evaluation script\footnote{\url{https://github.com/tuetschek/e2e-metrics}}.

\section{Results and Analysis}
Table~\ref{tab:test_results_e2e} and~\ref{tab:test_results_webnlg} display the results on the E2E and WebNLG test sets for models of the respective challenges and our own models\footnote{For an exact comparison, we recomputed the WebNLG~challenge results with the E2E evaluation script. They are usually 1-2~points below the scores reported by~\citet{WebNLGReport_Gardent17}.}.
Since the performance of neural models can vary considerably due to random parameter initialization and randomized training procedures~\cite{RandomVar_Reimers17}, we train ten models with different random seeds for each setting and report the average~(avg) and standard deviation~(SD).

On the E2E test set, our single best word- and character-based models reach comparable results to the best challenge submissions.
The word-based models achieve significantly higher BLEU and ROUGE-L scores than the character-based models\footnote{All tests for significance in this paper are conducted with Wilcoxon rank sum tests with Bonferroni correction at a p-level of~$0.05$.}.
On the WebNLG test set, the BLEU score of our best word-based model outperforms the best challenge submission by a small margin.
The character-based model achieves a significantly higher ROUGE-L score than the word-based model, whereas the BLEU score difference is not significant.
In the following, we analyze our models in more detail.

\subsection{Analysis of Within-Model Performance Differences}
The large performance span of the character-based models on the E2E dataset is due to a single outlier model; the second worst model scores 64.5~BLEU points.
The worst-scoring model had a lower accuracy of 91.8\% on the development set, whereas all other models scored above~92.2\%.
To gain more insight on what might constitute the large performance difference, we manually compared the generated texts for ten randomly selected inputs for each number of attributes (60 inputs in total) of the character-based model with the best and worst BLEU~score.
We found that the worst model makes many mistakes on inputs with three to five attributes, often adding, modifying or removing information, whereas the outputs are mostly correct for inputs with six attributes or more.
For these, the outputs of the model with the lowest BLEU score are occasionally even better than those of the best model, which often omits information (mainly concerning the attribute \textit{family friendly}).
We conclude that the large performance difference might be caused by automatic evaluation measures punishing additions more severely than omissions.

We also observe a large performance span for the WebNLG word-based models.
Here, we have two models that score exceptionally well with 57.4/58.4~BLEU points, whereas the remaining eight models only obtain BLEU scores in a range of~43.8-48.1.
Again, we observe that better models in terms of BLEU score obtain higher accuracies on the development set.
We manually compared the outputs of ten randomly chosen inputs for each number of input triples (75 inputs in total) for the model with the highest and lowest BLEU score.
In this case, we found that the large difference in the automatic evaluation measures seems justified: The low-scoring model often hallucinates information not present in the input and generally produces many ungrammatical texts, which is not the case for the best model.

\begin{table}[htb]
	\centering
 	\begin{tabular}{lrr}
 	 system & BLEU & ROUGE-L\\
 	\midrule
    \multicolumn{3}{c}{challenge} \\
    \midrule
 	baseline 	& 65.9 & 68.5 \\
 	\small{Thomson Reuters (np 3)}	& \textbf{68.1} & 69.3 \\
    \small{Thomson Reuters (np 4)}	& 67.4 & 69.8 \\
    \small{HarvardNLP \& H. Elder}
    & 67.4 & \textbf{70.8} \\
    \midrule 
    \multicolumn{3}{c}{own} \\
    \midrule
    word		& \resultSt{67.8}{0.8} & \resultSt{70.4}{0.6} \\
    character 	& \resultSt{64.6}{6.0} & \resultSt{67.9}{4.7} \\
    word (best on dev.) & 67.8 & 70.2 \\
 	char. (best on dev.) & 67.6 & 70.4 \\ 
     \end{tabular}
	\caption{E2E~test set results.
    Own results correspond to \resultSt{avg}{SD} of ten runs
    and single result of best models on the development set.}
    	\label{tab:test_results_e2e}
\end{table}

\begin{table}[htb]
\centering
     \begin{tabular}{lrr}
 	 system & BLEU & ROUGE-L \\
 	\midrule
    \multicolumn{3}{c}{challenge} \\
    \midrule
 	baseline 	& 32.1 & 43.3 \\
    \textsc{Melbourne} 	& 43.4 & \textbf{61.0} \\
    \textsc{Tilburg-SMT} & 43.1 & 58.0\\
    \textsc{UPF-FORGe}	& 37.5 & 58.8 \\
    \midrule
    \multicolumn{3}{c}{own} \\
    \midrule
 	word (best on dev.)  & \textbf{44.2} & 60.9 \\
 	char. (best on dev.) & 41.3 & 58.4 \\
    word		& \resultSt{37.0}{3.8} & \resultSt{56.3}{2.6} \\
    character 	&  \resultSt{39.7}{1.7} & \resultSt{58.4}{0.7}\\
     \end{tabular}
	\caption{WebNLG~test set results.
    Own results correspond to single best model on development set and \resultSt{avg}{SD} of ten runs.}
    \label{tab:test_results_webnlg}
\end{table}

\subsection{Automatic Evaluation of Human Texts}
To gain an impression of the expressiveness of the automatic evaluation scores for NLG, we computed the average scores that the human references would obtain.
Table~\ref{tab:human_as_reference} shows the BLEU and ROUGE-L development set scores when treating each human reference as prediction once and evaluating it against the remaining references, compared to the scores of the word-based and character-based models\footnote{For a fair comparison between human and model performance,
we randomly removed one reference for each instance in the models' evaluation to ensure the same average number of references. We excluded 55~WebNLG instances that had only one reference.}.
Strikingly, on the E2E development set, both model variants significantly outperform human texts by far with respect to both automatic evaluation measures.
While the human BLEU score is significantly higher than those of both systems on the WebNLG development set, there is no statistical difference between human and system ROUGE-L scores.
This further demonstrates the limited utility of BLEU and ROUGLE-L scores to evaluate NLG outputs, which was previously suggested by weak correlations of such scores with human judgments~\cite{NLGMetricsEvaluation_Scott_06,NLGMetricsEvaluation_Reiter_09,NLGEval_Novikova_17}.
Furthermore, the high scores on the E2E dataset imply that the models succeed in picking up patterns from the training data that transfer well to the similar development set, whereas human variation and creativity are punished by lexical overlap-based automatic evaluation scores.

\begin{table}[htb]
\begin{tabular}{lrrr}
    metric &human & word & char.\\
	\midrule
   	\multicolumn{4}{c}{E2E}\\
    \midrule
	BLEU 				& \resultSt{55.5}{0.7} & \resultSt{68.2}{1.4} & \resultSt{65.8}{2.6}\\ 
	ROUGE-L 			& \resultSt{62.0}{0.4} & \resultSt{72.1}{0.7} & \resultSt{69.8}{2.6}\\
    \midrule
    \multicolumn{4}{c}{WebNLG}\\
    \midrule
    BLEU 				&  \resultSt{48.3}{0.7} & \resultSt{40.6}{4.2} & \resultSt{43.7}{2.4}\\ 
	ROUGE-L 			&  \resultSt{62.4}{0.3} & \resultSt{58.5}{3.0} & \resultSt{63.1}{0.8}\\
\end{tabular}
	\caption{E2E and WebNLG development set results in the format \resultSt{avg}{SD}. Human results are averaged over using each human reference as prediction once.
    }
   \label{tab:human_as_reference}
\end{table}

\begin{table}[htb]
 	\resizebox{\linewidth}{!}{
 	\begin{tabular}{lrrrr}
        & 	\multicolumn{2}{c}{E2E} & \multicolumn{2}{c}{WebNLG}\\
            & word & char. & word & char. \\
            \midrule
            \multicolumn{5}{c}{content errors} \\
            \midrule
       info. dropped 	& 40.0 & 30.0 & 42.9 & 66.7 \\
       info. added 		& 0.0 & 0.0 & 6.7 & 1.9 \\
       info. modified	&  4.4 & 0.0 &  19.0 & 1.9 \\
       info. repeated	& 0.0 & 0.0 & 15.2 & 28.6\\
       \midrule
       \multicolumn{5}{c}{linguistic errors} \\
       \midrule
       punctuation errors 	& 5.6 & 5.6 & 8.6 & 3.8\\
       grammatical errors 	& 13.3 & 14.4 & 15.2 & 12.4\\
       spelling mistakes	& 0.0 & 0.0 & 9.5 & 5.7\\
       \midrule
        \multicolumn{5}{c}{overall correctness} \\
       \midrule
       content correct & 55.6 & 70.0 & 46.7 & 31.4\\
       language correct & 83.3 & 81.1 & 69.5 & 79.0\\
       all correct & 48.9 & 61.1 & 33.3 & 26.7\\
   \end{tabular}
 	}
 	\caption{Percentage of affected instances in manual error analysis of 15~randomly selected development set instances for each input length.}
     	\label{tab:manual_error_analysis}
 \end{table}
 
 \begin{table*}
\resizebox{\linewidth}{!}{
\begin{tabular}{p{2.1cm}rrrrrr}
    & \multicolumn{3}{c}{E2E} & \multicolumn{3}{c}{WebNLG}\\
    \midrule
   	& human & word & character & human & word & character\\
    \midrule
    unique sents.& \resultSt{866.3}{16.5} & \resultSt{203.5}{30.6} 	& \resultSt{366.8}{60.0} & \resultSt{1,185.0}{12.6} & \resultSt{603.7}{144.3} & \resultSt{875.4}{30.2}\\
    unique words 	& \resultSt{419.7}{16.7} & \resultSt{64.4}{2.3}& \resultSt{73.1}{7.2} & \resultSt{1447.3}{7.4} & \resultSt{620.3}{35.5} &  \resultSt{881.5}{26.0}\\
    word E	& \resultSt{6.5}{0.0} & \resultSt{5.1}{0.0} & \resultSt{5.5}{0.0} & \resultSt{7.1}{0.0} & \resultSt{6.3}{0.0} & \resultSt{6.6}{0.0} \\
    1-3-grams E & \resultSt{10.4}{0.0} & \resultSt{7.7}{0.1} & \resultSt{8.2}{0.1} & \resultSt{11.6}{0.0} & \resultSt{10.1}{0.1} & \resultSt{10.5}{0.1}\\
    \% new texts & \resultSt{99.7}{0.2} & \resultSt{98.2}{0.3} & \resultSt{98.8}{0.2} & \resultSt{91.1}{0.3} & \resultSt{69.8}{4.8} & \resultSt{87.5}{0.6}\\
    \% new sents.& \resultSt{85.1}{1.1} & \resultSt{61.8}{6.4}& \resultSt{71.4}{4.7} & \resultSt{87.4}{0.4} & \resultSt{57.2}{5.8}& \resultSt{82.1}{1.2}
\end{tabular}
}
\caption{Linguistic diversity of development set references and generated texts as \resultSt{avg}{SD}. \lq \%~new\rq ~denotes the share of generated texts or sentences that do not appear in training references.
    Higher indicates more diversity for all measures.}
\label{tab:results_diversity}
\end{table*}

\subsection{Manual Error Analysis}
Since the expressiveness of automatic evaluation measures for NLG is limited, as shown in the previous subsection, we performed a manual error analysis on inputs of each length.
We define the input length as the number of input attributes for the E2E dataset, ranging from three to eight, and number of input triples for the WebNLG dataset, ranging from one to seven. 
We randomly selected 15 development instances for each input length, resulting in a total of 90~annotated E2E instances and 105~WebNLG instances.

One annotator (one of the authors of this paper) manually assessed the outputs of the models that obtained the best development set BLEU score as summarized in Table~\ref{tab:manual_error_analysis}\footnote{Although multiple annotators could increase the reliability of these results, the annotator reported that the task was very straightforward. We do not expect marking content and linguistic errors to lead to annotator disagreements, with the exception of accidentally missed errors.}.
As we can see from the bottom part of the table, all models struggle more with getting the content right than with producing linguistically correct texts; 70-80\%~of the texts generated by all models are completely correct linguistically.

Comparing the two datasets, we again observe that the WebNLG dataset is much more challenging than the E2E dataset, especially with respect to correctly verbalizing the content.
This can be attributed to the increased diversity of the inputs and texts and to the limited availability of training data for this dataset (cf.~Table~\ref{tab:dataset}).
Moreover, spelling mistakes only appeared in WebNLG texts, mainly concerning omissions of accents or umlauts.
This also indicates that there is too few and noisy data for the models to learn the correct spelling of all words.
Notably, we did not observe any non-words generated by the character-based models.

The most frequent content error in both datasets concerns omission of information.
For the E2E dataset, the \textit{family friendly} attribute is most frequently dropped by both model types, indicating that the verbalization of this boolean attribute is more difficult to learn than other attributes, whose values mostly appear verbatim in the text.
Information modification of the word-based model is mainly due to confusing \textit{English} with \textit{Italian} food.
Information addition and repetition only occur in the WebNLG dataset.
The latter is an especially frequent problem of the character-based model, affecting more than a quarter of all texts.

In comparison, character-based models reproduce
the content more faithfully on the E2E dataset while offering the same level of linguistic quality as word-based models, leading to more correct outputs overall.
On the WebNLG dataset, the word-based model is more faithful to the inputs, probably because of the effective delexicalization strategy, whereas the character-based model errs less on the linguistic side.
Overall, the word-based model yields more correct texts, stressing the importance of delexicalization and data normalization in low resource settings.

\subsection{Automatic Evaluation of Output Diversity}
While correctness is a necessity in NLG, in many settings it is not sufficient.
Often, variation of the generated texts is crucial to avoid repetitive and unnatural outputs.
Table~\ref{tab:results_diversity} shows automatically computed statistics on the diversity of the generated texts of both models and human texts and on the overlap of the (generated) texts with the training set.
We measure diversity by the number of unique sentences and words in all development set references and generated texts, as done e.g. by~\citet{LM4ImageCaptioning_Devlin15}.
Additionally, we report the Shannon text entropy as measure of the amount of variation in the texts following~\citep{PersonalityVariationNeuralNLG_Oraby18}.
We compute the text entropy~E for words (unigrams) and uni-, bi-, and trigrams as follows:

\begin{equation}
 E = - \sum _{w \in V} \frac{f(w)}{\text{total}} * \log_2{\frac{f(w)}{\text{total}}}
\end{equation}

where $V$ is the set of all word types or uni-, bi- and trigrams, $f$ denotes frequency and total is the token count or total number of uni-, bi- and trigrams in the texts, respectively. 

To measure the extent by which the models generalize beyond plugging in restaurant or other entity names into templates extracted from the training data, we compute the results on the delexicalized outputs of the word-based models and delexicalize the character-based models' outputs.
For the human scores, we generate $n$~artificial prediction files, treating each $n$-th reference (42~for E2E, 8~for WebNLG) as reference, apply delexicalization, and average the scores for the $n$~files.

\begin{figure*}
    \textbf{Template~1}: \textcolor{template1color}{\uwave{\textsc{name} is a [\textsc{family-friendly}] \textsc{eattype}} which serves [\textsc{food}] food [in the \textsc{price range} price range]. \textcolor{template1color}{\uwave{[It has a \textsc{rating} rating] [and is located in the \textsc{area} area[, near \textsc{near}]].}} [It is not \textsc{family-friendly}}.]\\
    \textbf{Example:} \textsc{name} is a family-friendly coffee shop which serves Chinese food in the low price range. It has a high customer rating and is located in the city centre area, near \textsc{near}.
\vspace{2mm}

    \textbf{Template~2}: \textcolor{template2color}{\underline{The [\textsc{family-friendly}] \textsc{eattype name}}} serves [\textsc{food}] food [in the \textsc{price range} price range]. \textcolor{template2color}{\underline{[It is located in the \textsc{area} area[, near \textsc{near}].] [It has a \textsc{rating} rating.]}} [It is not \textsc{family-friendly}.]\\
   \textbf{Example:} The family-friendly coffee shop \textsc{name} serves Chinese food in the low price range. It is located in the city centre area, near \textsc{near}. It has a high customer rating.
   
   \vspace{2mm}
   \textbf{Learned combinations of Template 1 and 2:}\\
   $\bullet$ \textcolor{template1color}{\uwave{\textsc{name} is a restaurant}} which serves English food in the moderate price range. \textcolor{template2color}{\underline{It is located in the city} \underline{centre area, near \textsc{near}. It has a customer rating of 1 out of 5.}} It is not family friendly.\\
   $\bullet$ \textcolor{template2color}{\underline{The family-friendly pub \textsc{name}}} serves Indian food in the low price range. \textcolor{template1color}{\uwave{It has a customer rating of 5 out of 5 and is located in the riverside area, near \textsc{near}.}}
    \caption{Templates used for synthetic training data generation, parts in brackets are realized only if the input contains the corresponding attribute. Learned combinations are two template combinations produced by a model trained on data generated from both templates.}
    \label{fig:templates}
\end{figure*}

On both datasets, our systems produce significantly less varied outputs and reproduce more texts and sentences from the training data than the human texts.
Interestingly, however, the character-based models generate significantly more unique sentences and copy significantly less from the training data than the word-based models, which copy about 40\%~of their generated sentences from the training data.

\section{Generalizing from Templates}
In search for empirical evidence that neural models are able to surpass the structures they were trained on, we train Seq2Seq models with synthetic training data created by templates.
This enables us to control the variation in the training data and identify novel generations of the model (if any).
We investigate two questions: (1) Do the neural NLG models indeed accurately learn the templates from the training data? (2) Do they learn to combine the training templates to produce more varied outputs than seen during training?

We generate synthetic training data based on two templates.
Template~1 corresponds to \textsc{UKP-TUDA}'s submission to the E2E challenge\footnote{\url{https://github.com/UKPLab/e2e-nlg-challenge-2017/blob/master/components/template-baseline.py}}, where the order of describing the input information is fixed. Specifically, the restaurant's customer rating is always mentioned before its location. For Template~2, we change the the beginning of the template and switch the order of mentioning the rating and location of the restaurant as shown in Figure~\ref{fig:templates}.
Potential combinations of the two templates are to combine the beginning of Template~1 with the ordering of rating and area of Template~2 or vice versa.
We generate a single reference text for all 2261~training inputs of the E2E dataset where the \textsc{name} and \textsc{eattype} attribute are present as these are the two obligatory attributes for the templates.
We train word-based models on training data generated with Template~1, Template~2 and the concatenation of the training data from Template~1 and~2. To keep the amount of training data equal in all experiments, we once repeat the training corpus generated only with Template~1 or Template~2.
The hyperparameters for the three models can be found in the appendix.
%

\begin{table}[hbt]
  \begin{tabular}{l|rrrrr}
  						& c@1 & c@2 & c@5 & c@30\\
   \midrule
   template~1			&0.8 & 0.8 & 0.9 & 1.7\\
   template~2			&1.0 & 1.2 & 1.3 & 1.9\\
   template~1+2			&0.9 & 1.6 & 2.2 & 3.3\\
   + reranker 			&0.9 & 1.9 & 2.7 & 3.3 \\
  \end{tabular}
      \caption{Manual evaluation of generated texts for 10~random test instances of a word-based model trained with synthetic training data from two templates. c@n: avg. number of correct texts (with respect to content and language) among the top~n hypotheses.}
      \label{tab:results_synth}
\end{table}

Table~\ref{tab:results_synth} shows our manual evaluation of the top 30~hypotheses for 10~random E2E test inputs generated by models trained with data synthesized from the two templates.
As is evident from the first two rows, all models learned to generalize from the training data to produce correct texts for novel inputs consisting of unseen combinations of input attributes.
It was verified in the manual evaluation that 100\% of the texts generated by models trained on a single template adhered to this template.
Yet, the picture is a bit different for the model trained on data generated by both templates.
While the top two hypotheses are equally distributed between adhering to Template~1 and Template~2, more than 5\% among the lower-ranked hypotheses constitute a template combination such as the example shown in the bottom part of Figure~\ref{fig:templates}.
For 60\% of the examined inputs, there was at least one such hypothesis resulting from template combination, of which two thirds were actually correct verbalizations of the input.

Since we found that the models frequently ranked correct hypotheses below hypotheses with content errors, we implemented a simple rule-based reranker based on verbatim matches of attribute values.
The reranker assigns an error point to each omission and addition of an attribute value.
As can be seen in the final row of Table~\ref{tab:results_synth}, this simple reranker successfully places correct hypotheses higher up in the ranking, improving the practical usability of the generation model by now offering almost three correct variants for each input among the top five hypotheses on average.

\section{Conclusion}
We compared word-based and character-based Seq2Seq models for data-to-text NLG on two datasets and analyzed their output diversity.
Our main findings are as follows:
Overall, Seq2Seq models can learn to verbalize structured inputs in a decent way; their success depends on the extent of the domain and available (clean) training data.

Second, in a comparison with texts produced by humans, we saw that neural NLG models can even surpass human performance in terms of automatic evaluation measures.
On the one hand, this unveils the ability of the models to extract general patterns from the training data that approximate many reference texts, but on the other hand also once more stresses the limited utility of such measures to evaluate NLG systems.

Third, in light of the multi-faceted analysis we performed, it is difficult to draw a general conclusion on whether word- or character-based processing is more useful for data-to-text generation.
Both models yielded comparable results with respect to automatic evaluation measures.
In the manual error analysis, the character-based model performed better on the E2E dataset, whereas the word-based model generated more correct outputs on the WebNLG dataset. Character-based models were found to have a significantly higher output diversity.

Finally, in a controlled experiment with word-based Seq2Seq models trained on data synthesized from templates, we showed the capability of such models to perfectly reproduce the templates they were trained on. More importantly, models trained on two templates could generalize beyond their training data and come up with novel texts.
In future work, we would like to extend this line of research and train more model variants on a higher number of templates.

\bibliography{NLG,dialogSystems,NN,NNCM}

\begin{thebibliography}{58}
\expandafter\ifx\csname natexlab\endcsname\relax\def\natexlab#1{#1}\fi

\bibitem[{Agarwal and Dymetman(2017)}]{Char2Char4E2E_Shubham_2017}
Shubham Agarwal and Marc Dymetman. 2017.
\newblock \href {http://www.aclweb.org/anthology/W17-3619} {A surprisingly
  effective out-of-the-box char2char model on the e2e nlg challenge dataset}.
\newblock In \emph{Proceedings of the 18th Annual SIGdial Meeting on Discourse
  and Dialogue}, pages 158--163, Saarbr{\"u}cken, Germany.

\bibitem[{Angeli et~al.(2010)Angeli, Liang, and
  Klein}]{RobocubWeatherGov_Angeli10}
Gabor Angeli, Percy Liang, and Dan Klein. 2010.
\newblock \href {http://dl.acm.org/citation.cfm?id=1870658.1870707} {A simple
  domain-independent probabilistic approach to generation}.
\newblock In \emph{Proceedings of the 2010 Conference on Empirical Methods in
  Natural Language Processing}, EMNLP '10, pages 502--512, Stroudsburg, PA,
  USA.

\bibitem[{Bahdanau et~al.(2014)Bahdanau, Cho, and
  Bengio}]{Attention_Bahdanau14}
Dzmitry Bahdanau, Kyunghyun Cho, and Yoshua Bengio. 2014.
\newblock {Neural Machine Translation by Jointly Learning to Align and
  Translate}.
\newblock \emph{arXiv e-prints}, abs/1409.0473.

\bibitem[{Bawden(2017)}]{StyleMT_Bawden17}
Rachel Bawden. 2017.
\newblock \href {http://aclweb.org/anthology/D17-1265} {Machine translation,
  it's a question of style, innit? the case of english tag questions}.
\newblock In \emph{Proceedings of the 2017 Conference on Empirical Methods in
  Natural Language Processing, {EMNLP} 2017}, pages 2507--2512, Copenhagen,
  Denmark.

\bibitem[{Bird et~al.(2009)Bird, Klein, and Loper}]{NLTK_Bird09:NLP}
Steven Bird, Ewan Klein, and Edward Loper. 2009.
\newblock \emph{{Natural Language Processing with Python}}, 1st edition.
\newblock O'Reilly Media, Inc.

\bibitem[{Cheyer and Guzzoni(2006)}]{NLGTemplateDialog_Cheyer06}
Adam Cheyer and Didier Guzzoni. 2006.
\newblock Method and apparatus for building an intelligent automated assistant.
\newblock Patent US 11/518,292 (Patent pending).

\bibitem[{Cho et~al.(2014)Cho, Van~Merri{\"{e}}nboer, G{\"{u}}l{\c c}ehre,
  Bahdanau, Bougares, Schwenk, and Bengio}]{Seq2Seq_Cho14}
Kyunghyun Cho, Bart Van~Merri{\"{e}}nboer, {\c C}ağlar G{\"{u}}l{\c c}ehre,
  Dzmitry Bahdanau, Fethi Bougares, Holger Schwenk, and Yoshua Bengio. 2014.
\newblock \href {http://www.aclweb.org/anthology/D14-1179} {Learning phrase
  representations using rnn encoder--decoder for statistical machine
  translation}.
\newblock In \emph{Proceedings of the 2014 Conference on Empirical Methods in
  Natural Language Processing (EMNLP)}, pages 1724--1734, Doha, Qatar.

\bibitem[{Chung et~al.(2016)Chung, Cho, and Bengio}]{Bpe2Char_Chung16}
Junyoung Chung, Kyunghyun Cho, and Yoshua Bengio. 2016.
\newblock \href {http://aclweb.org/anthology/P/P16/P16-1160.pdf} {A
  character-level decoder without explicit segmentation for neural machine
  translation}.
\newblock In \emph{Proceedings of the 54th Annual Meeting of the Association
  for Computational Linguistics, {ACL} 2016, Volume 1: Long Papers}, pages
  1693--1703, Berlin, Germany.

\bibitem[{Devlin et~al.(2015)Devlin, Cheng, Fang, Gupta, Deng, He, Zweig, and
  Mitchell}]{LM4ImageCaptioning_Devlin15}
Jacob Devlin, Hao Cheng, Hao Fang, Saurabh Gupta, Li~Deng, Xiaodong He,
  Geoffrey Zweig, and Margaret Mitchell. 2015.
\newblock Language models for image captioning: The quirks and what works.
\newblock In \emph{Proceedings of the 53rd Annual Meeting of the Association
  for Computational Linguistics and the 7th International Joint Conference on
  Natural Language Processing of the Asian Federation of Natural Language
  Processing, {ACL} 2015, Volume 2: Short Papers}, pages 100--105, Beijing,
  China.

\bibitem[{Du{\v{s}}ek and Jurc{\'{\i}}cek(2016)}]{Seq2Tree_Dusek16}
Ond{\v{r}}ej Du{\v{s}}ek and Filip Jurc{\'{\i}}cek. 2016.
\newblock \href {http://aclweb.org/anthology/P/P16/P16-2008.pdf}
  {Sequence-to-sequence generation for spoken dialogue via deep syntax trees
  and strings}.
\newblock In \emph{Proceedings of the 54th Annual Meeting of the Association
  for Computational Linguistics, {ACL} 2016, Volume 2: Short Papers}, pages
  41--51, Berlin, Germany.

\bibitem[{Ficler and Goldberg(2017)}]{ControlingLinguisticStyleNLG_Ficler_17}
Jessica Ficler and Yoav Goldberg. 2017.
\newblock \href {http://arxiv.org/abs/1707.02633} {Controlling linguistic style
  aspects in neural language generation}.
\newblock \emph{CoRR}, abs/1707.02633.

\bibitem[{Gardent et~al.(2017{\natexlab{a}})Gardent, Shimorina, Narayan, and
  Perez{-}Beltrachini}]{WebNLGTrainingCorporaNLGMicroPlanners_Gardent17}
Claire Gardent, Anastasia Shimorina, Shashi Narayan, and Laura
  Perez{-}Beltrachini. 2017{\natexlab{a}}.
\newblock \href {https://doi.org/10.18653/v1/P17-1017} {Creating training
  corpora for {NLG} micro-planners}.
\newblock In \emph{Proceedings of the 55th Annual Meeting of the Association
  for Computational Linguistics, {ACL} 2017, Volume 1: Long Papers}, pages
  179--188, Vancouver, Canada.

\bibitem[{Gardent et~al.(2017{\natexlab{b}})Gardent, Shimorina, Narayan, and
  Perez-Beltrachini}]{WebNLGReport_Gardent17}
Claire Gardent, Anastasia Shimorina, Shashi Narayan, and Laura
  Perez-Beltrachini. 2017{\natexlab{b}}.
\newblock \href {http://aclweb.org/anthology/W17-3518} {{The WebNLG Challenge:
  Generating Text from RDF Data}}.
\newblock In \emph{Proceedings of the 10th International Conference on Natural
  Language Generation}, pages 124--133, Santiago de Compostela, Spain.

\bibitem[{Gatt and Krahmer(2018)}]{NLGSurvey_Gatt18}
Albert Gatt and Emiel Krahmer. 2018.
\newblock \href {http://arxiv.org/abs/https://arxiv.org/abs/1703.09902} {Survey
  of the state of the art in natural language generation: Core tasks,
  applications and evaluation}.
\newblock \emph{Journal of Artificial Intelligence Research (JAIR)},
  61:65--170.

\bibitem[{Goyal et~al.(2016)Goyal, Dymetman, and
  Gaussier}]{CharNLG_Finite_State_Goyal_16}
Raghav Goyal, Marc Dymetman, and {\'{E}}ric Gaussier. 2016.
\newblock \href {http://aclweb.org/anthology/C/C16/C16-1103.pdf} {Natural
  language generation through character-based rnns with finite-state prior
  knowledge}.
\newblock In \emph{Proceedings of the 26th International Conference on
  Computational Linguistics, {COLING} 2016, Technical Papers}, pages
  1083--1092, Osaka, Japan.

\bibitem[{Graves(2013)}]{Seq2Seq_Graves13}
Alex Graves. 2013.
\newblock \href {http://arxiv.org/abs/1308.0850} {Generating sequences with
  recurrent neural networks}.
\newblock \emph{CoRR}, abs/1308.0850.

\bibitem[{Gu et~al.(2016)Gu, Lu, Li, and
  Li}]{CopyingInSequenceToSequence_IncorporatingGu16}
Jiatao Gu, Zhengdong Lu, Hang Li, and Victor O.~K. Li. 2016.
\newblock \href {http://aclweb.org/anthology/P/P16/P16-1154.pdf} {Incorporating
  copying mechanism in sequence-to-sequence learning}.
\newblock In \emph{Proceedings of the 54th Annual Meeting of the Association
  for Computational Linguistics, {ACL} 2016, Volume 1: Long Papers}, pages
  1631--1640, Berlin, Germany.

\bibitem[{Herzig et~al.(2017)Herzig, Shmueli{-}Scheuer, Sandbank, and
  Konopnicki}]{ResponseGenerationCustomerSerivePersonality_Herzig17}
Jonathan Herzig, Michal Shmueli{-}Scheuer, Tommy Sandbank, and David
  Konopnicki. 2017.
\newblock Neural response generation for customer service based on personality
  traits.
\newblock In \emph{Proceedings of the 10th International Conference on Natural
  Language Generation, {INLG} 2017}, pages 252--256, Santiago de Compostela,
  Spain.

\bibitem[{Hochreiter and Schmidhuber(1997)}]{HochreiterLSTM_97}
Sepp Hochreiter and J\"{u}rgen Schmidhuber. 1997.
\newblock {Long Short-Term Memory}.
\newblock \emph{Neural Computation}, 9(8).

\bibitem[{Kiddon et~al.(2016)Kiddon, Zettlemoyer, and
  Choi}]{GloballyCoherent_NLG_Kiddon16}
Chlo{\'{e}} Kiddon, Luke Zettlemoyer, and Yejin Choi. 2016.
\newblock Globally coherent text generation with neural checklist models.
\newblock In \emph{{Proceedings of the 2016 Conference on Empirical Methods in
  Natural Language Processing, EMNLP 2016}}, pages 329--339, Austin, TX, USA.

\bibitem[{Kingma and Ba(2015)}]{Adam_Kingma15}
Diederik Kingma and Jimmy Ba. 2015.
\newblock {Adam: {A} Method for Stochastic Optimization}.
\newblock In \emph{International Conference on Learning Representations}, San
  Diego, CA, USA.

\bibitem[{Klein et~al.(2017)Klein, Kim, Deng, Senellart, and
  Rush}]{OpenNMT_Klein17}
Guillaume Klein, Yoon Kim, Yuntian Deng, Jean Senellart, and Alexander~M. Rush.
  2017.
\newblock \href {http://arxiv.org/abs/1701.02810} {Opennmt: Open-source toolkit
  for neural machine translation}.
\newblock \emph{CoRR}, abs/1701.02810.

\bibitem[{Koehn(2017)}]{NMTChapter_Koehn17}
Philipp Koehn. 2017.
\newblock \href {http://arxiv.org/abs/1709.07809} {Neural machine translation}.
\newblock \emph{CoRR}, abs/1709.07809.

\bibitem[{Lee et~al.(2017)Lee, Cho, and Hofmann}]{Char2CharNMT_Lee17}
Jason Lee, Kyunghyun Cho, and Thomas Hofmann. 2017.
\newblock \href {https://transacl.org/ojs/index.php/tacl/article/view/1051}
  {Fully character-level neural machine translation without explicit
  segmentation}.
\newblock \emph{{TACL}}, 5:365--378.

\bibitem[{Li et~al.(2016)Li, Galley, Brockett, Spithourakis, Gao, and
  Dolan}]{Persona_Li16}
Jiwei Li, Michel Galley, Chris Brockett, Georgios Spithourakis, Jianfeng Gao,
  and William~B. Dolan. 2016.
\newblock {A Persona-Based Neural Conversation Model}.
\newblock In \emph{Proceedings of the 54th Annual Meeting of the Association
  for Computational Linguistics, {ACL} 2016, Volume 1: Long Papers}, pages
  994--1003, Berlin, Germany.

\bibitem[{Lin(2004)}]{ROUGE-L_Lin04}
Chin-Yew Lin. 2004.
\newblock {ROUGE: A Package for Automatic Evaluation of summaries}.
\newblock In \emph{Proceedings of the ACL workshop on Text Summarization
  Branches Out}, pages 74--81, Barcelona, Spain.

\bibitem[{Lin and Walker(2011)}]{FilmCharStage_Lin11}
Grace Lin and Marilyn Walker. 2011.
\newblock All the world’s a stage: Learning character models from film.
\newblock In \emph{Proceedings of the Seventh AIIDE Conference}, pages 46--52,
  Palo Alto, CA, USA.

\bibitem[{Ling et~al.(2015)Ling, Trancoso, Dyer, and
  Black}]{CharBasedNMT_Ling15}
Wang Ling, Isabel Trancoso, Chris Dyer, and Alan Black. 2015.
\newblock \href {http://arxiv.org/abs/1511.04586} {Character-based neural
  machine translation}.
\newblock \emph{CoRR}, abs/1511.04586.

\bibitem[{Lipton et~al.(2015)Lipton, Vikram, and
  McAuley}]{CharacterLevelGenerative_Lipton15}
Zachary~Chase Lipton, Sharad Vikram, and Julian McAuley. 2015.
\newblock \href {http://arxiv.org/abs/1511.03683} {Capturing meaning in product
  reviews with character-level generative text models}.
\newblock \emph{CoRR}, abs/1511.03683.

\bibitem[{Luong et~al.(2015)Luong, Pham, and Manning}]{Attention_Luong15}
Thang Luong, Hieu Pham, and Christopher Manning. 2015.
\newblock Effective approaches to attention-based neural machine translation.
\newblock In \emph{Proceedings of the 2015 Conference on Empirical Methods in
  Natural Language Processing, {EMNLP} 2015}, pages 1412--1421, Lisbon,
  Portugal.

\bibitem[{Mairesse and Young(2014)}]{NLGCorpusBasedFactoredLM_Mairesse14}
Fran\c{c}ois Mairesse and Steve~J. Young. 2014.
\newblock \href {https://doi.org/10.1162/COLI\_a\_00199} {Stochastic language
  generation in dialogue using factored language models}.
\newblock \emph{Computational Linguistics}, 40(4):763--799.

\bibitem[{Mairesse and Walker(2008)}]{Big5PersonalityTraits_Mariesse08}
Francois Mairesse and Marilyn Walker. 2008.
\newblock {T}rainable {G}eneration of {B}ig-{F}ive {P}ersonality {S}tyles
  through {D}ata-driven {P}arameter {E}stimation.
\newblock In \emph{{Proceedings of the 46th Annual Meeting of the Association
  for Computational Linguistics (ACL 2008)}}, pages 165--173, Columbus, OH,
  USA.

\bibitem[{Mei et~al.(2016)Mei, Bansal, and Walter}]{SelectiveNLG_Mei_16}
Hongyuan Mei, Mohit Bansal, and Matthew Walter. 2016.
\newblock What to talk about and how? selective generation using lstms with
  coarse-to-fine alignment.
\newblock In \emph{Proceedings of the Conference of the North American Chapter
  of the Association for Computational Linguistics — Human Language
  Technologies (NAACL HLT)}, pages 720--730, San Diego, CA, USA.

\bibitem[{Mirkovic et~al.(2006)Mirkovic, Cavedon, Purver, Ratiu, Scheideck,
  Weng, Zhang, and Xu}]{NLGTemplateDialog_Mirkovic06}
Danilo Mirkovic, Lawrence Cavedon, Matthew Purver, Florin Ratiu, Tobias
  Scheideck, Fuliang Weng, Qi~Zhang, and Kui Xu. 2006.
\newblock \href {https://www.google.com/patents/US20060271364} {Dialogue
  management using scripts and combined confidence scores}.
\newblock US Patent App. 11/298,765.

\bibitem[{Niu et~al.(2017)Niu, Martindale, and Carpuat}]{FormalStyleMT_Niu17}
Xing Niu, Marianna Martindale, and Marine Carpuat. 2017.
\newblock \href {https://aclanthology.info/papers/D17-1299/d17-1299} {A study
  of style in machine translation: Controlling the formality of machine
  translation output}.
\newblock In \emph{Proceedings of the 2017 Conference on Empirical Methods in
  Natural Language Processing, {EMNLP} 2017}, pages 2814--2819, Copenhagen,
  Denmark.

\bibitem[{Novikova et~al.(2017{\natexlab{a}})Novikova, Du\v{s}ek, Curry, and
  Rieser}]{NLGEval_Novikova_17}
Jekaterina Novikova, Ond\v{r}ej Du\v{s}ek, Amanda~Cercas Curry, and Verena
  Rieser. 2017{\natexlab{a}}.
\newblock \href {http://aclanthology.info/papers/D17-1237/d17-1237} {Why we
  need new evaluation metrics for {NLG}}.
\newblock In \emph{Proceedings of the 2017 Conference on Empirical Methods in
  Natural Language Processing, {EMNLP} 2017}, pages 2231--2242, Copenhagen,
  Denmark.

\bibitem[{Novikova et~al.(2017{\natexlab{b}})Novikova, Du\v{s}ek, and
  Rieser}]{E2E_Novikova_2017}
Jekaterina Novikova, Ond\v{r}ej Du\v{s}ek, and Verena Rieser.
  2017{\natexlab{b}}.
\newblock \href {http://www.aclweb.org/anthology/W17-3625} {The e2e dataset:
  New challenges for end-to-end generation}.
\newblock In \emph{Proceedings of the 18th Annual SIGdial Meeting on Discourse
  and Dialogue}, pages 201--206, Saarbr{\"u}cken, Germany.

\bibitem[{Oh and Rudnicky(2000)}]{NLGCorpusBasedOvergeneration_Oh00}
Alice Oh and Alexander Rudnicky. 2000.
\newblock \href {https://doi.org/10.3115/1117562.1117568} {Stochastic language
  generation for spoken dialogue systems}.
\newblock In \emph{Proceedings of the 2000 ANLP/NAACL Workshop on
  Conversational Systems - Volume 3}, ANLP/NAACL-ConvSyst '00, pages 27--32,
  Stroudsburg, PA, USA.

\bibitem[{Oraby et~al.(2018{\natexlab{a}})Oraby, Reed, S., Tandon, and
  Walker}]{MultiVoice4PersonalityNeuralDialog_Oraby18}
Shereen Oraby, Lena Reed, Sharath~T. S., Shubhangi Tandon, and Marilyn~A.
  Walker. 2018{\natexlab{a}}.
\newblock Neural multivoice models for expressing novel personalities in
  dialog.
\newblock In \emph{Interspeech}, pages 3057--3061, Hyderabad, India. {ISCA}.

\bibitem[{Oraby et~al.(2018{\natexlab{b}})Oraby, Reed, Tandon, S., Lukin, and
  Walker}]{PersonalityVariationNeuralNLG_Oraby18}
Shereen Oraby, Lena Reed, Shubhangi Tandon, Sharath~T. S., Stephanie Lukin, and
  Marilyn Walker. 2018{\natexlab{b}}.
\newblock \href {https://aclanthology.info/papers/W18-5019/w18-5019}
  {Controlling personality-based stylistic variation with neural natural
  language generators}.
\newblock In \emph{Proceedings of the 19th Annual SIGdial Meeting on Discourse
  and Dialogue}, pages 180--190, Melbourne, Australia.

\bibitem[{Papineni et~al.(2002)Papineni, Roukos, Ward, and
  Zhu}]{BLEU_Papineni02}
Kishore Papineni, Salim Roukos, Todd Ward, and Wei-Jing Zhu. 2002.
\newblock Bleu: A method for automatic evaluation of machine translation.
\newblock In \emph{Proceedings of the 40th Annual Meeting on Association for
  Computational Linguistics}, ACL '02, pages 311--318, Stroudsburg, PA, USA.

\bibitem[{Reimers and Gurevych(2017)}]{RandomVar_Reimers17}
Nils Reimers and Iryna Gurevych. 2017.
\newblock \href {http://aclweb.org/anthology/D17-1035} {Reporting score
  distributions makes a difference: Performance study of lstm-networks for
  sequence tagging}.
\newblock In \emph{Proceedings of the 2017 Conference on Empirical Methods in
  Natural Language Processing}, pages 338--348, Copenhagen, Denmark.

\bibitem[{Reiter and Belz(2009)}]{NLGMetricsEvaluation_Reiter_09}
Ehud Reiter and Anja Belz. 2009.
\newblock \href {https://doi.org/10.1162/coli.2009.35.4.35405} {An
  investigation into the validity of some metrics for automatically evaluating
  natural language generation systems}.
\newblock \emph{Computational Linguistics}, 35(4):529--558.

\bibitem[{Robbins and Monro(1951)}]{SGD_Robbins51}
Herbert Robbins and Sutton Monro. 1951.
\newblock A stochastic approximation method.
\newblock \emph{Annals of Mathematical Statistics}, 22:400--407.

\bibitem[{Scott and Moore(2006)}]{NLGMetricsEvaluation_Scott_06}
Donia Scott and Johanna Moore. 2006.
\newblock {An NLG evaluation competition? Eight Reasons to Be Cautious}.
\newblock In \emph{Proceedings of the Fourth International Natural Language
  Generation Conference, {INLG} 2006, Special Session on Sharing Data and
  Comparative Evaluations}, Sydney, Australia.

\bibitem[{See et~al.(2017)See, Manning, and
  Liu}]{GetToThePointSummarization_See17}
Abigail See, Christopher Manning, and Peter Liu. 2017.
\newblock \href {https://arxiv.org/abs/1704.04368} {Get to the point:
  Summarization with pointer-generator networks}.
\newblock In \emph{Association for Computational Linguistics}.

\bibitem[{Sennrich et~al.(2016)Sennrich, Haddow, and
  Birch}]{SubwordUnits_Sennrich16}
Rico Sennrich, Barry Haddow, and Alexandra Birch. 2016.
\newblock \href {http://aclweb.org/anthology/P/P16/P16-1162.pdf} {Neural
  machine translation of rare words with subword units}.
\newblock In \emph{Proceedings of the 54th Annual Meeting of the Association
  for Computational Linguistics, {ACL} 2016, August 7-12, 2016, Berlin,
  Germany, Volume 1: Long Papers}.

\bibitem[{Stent et~al.(2005)Stent, Marge, and
  Singhai}]{NLGEvalVariation_Stent05}
Amanda Stent, Matthew Marge, and Mohit Singhai. 2005.
\newblock Evaluating evaluation methods for generation in the presence of
  variation.
\newblock In \emph{Computational Linguistics and Intelligent Text Processing},
  pages 341--351, Berlin, Heidelberg. Springer.

\bibitem[{Sutskever et~al.(2011)Sutskever, Martens, and
  Hinton}]{GeneratingTextsRNN_Sutskever11}
Ilya Sutskever, James Martens, and Geoffrey Hinton. 2011.
\newblock Generating text with recurrent neural networks.
\newblock In \emph{Proceedings of the 28th International Conference on Machine
  Learning, {ICML} 2011}, pages 1017--1024, Bellevue, WA, USA.

\bibitem[{Sutskever et~al.(2014)Sutskever, Vinyals, and
  Le}]{Seq2Se2_Sutskever14}
Ilya Sutskever, Oriol Vinyals, and Quoc~V. Le. 2014.
\newblock {Sequence to Sequence Learning with Neural Networks}.
\newblock In \emph{Proceedings of the 27th International Conference on Neural
  Information Processing Systems}, NIPS'14, Cambridge, Massachusetts, USA. MIT
  Press.

\bibitem[{Vinyals and Le(2015)}]{NeuralConvModel_Vinyals15}
Oriol Vinyals and Quoc~V. Le. 2015.
\newblock A neural conversational model.
\newblock In \emph{Proceedings of the International Conference on Machine
  Learning, Deep Learning Workshop}, Lille, France.

\bibitem[{Walker et~al.(2011)Walker, Grant, Sawyer, Lin, Wardrip-Fruin, and
  Buell}]{FilmCharacterNLG_Walker11}
Marilyn~A. Walker, Ricky Grant, Jennifer Sawyer, Grace~I. Lin, Noah
  Wardrip-Fruin, and Michael Buell. 2011.
\newblock {Perceived or Not Perceived: Film Character Models for Expressive
  NLG.}
\newblock In \emph{ICIDS}, volume 7069 of \emph{Lecture Notes in Computer
  Science}. Springer.

\bibitem[{Wen et~al.(2015{\natexlab{a}})Wen, Gasic, Kim, Mrksic, Su, Vandyke,
  and Young}]{StochasticLanguageGenerationRNNCNNReranking_Wen15}
Tsung{-}Hsien Wen, Milica Gasic, Dongho Kim, Nikola Mrksic, Pei{-}hao Su, David
  Vandyke, and Steve~J. Young. 2015{\natexlab{a}}.
\newblock \href {http://arxiv.org/abs/1508.01755} {Stochastic language
  generation in dialogue using recurrent neural networks with convolutional
  sentence reranking}.
\newblock \emph{CoRR}, abs/1508.01755.

\bibitem[{Wen et~al.(2016)Wen, Gasic, Mrksic, Rojas{-}Barahona, Su, Vandyke,
  and Young}]{MultiDomainSC_LSTM_Wen_16}
Tsung{-}Hsien Wen, Milica Gasic, Nikola Mrksic, Lina~Maria Rojas{-}Barahona,
  Pei{-}Hao Su, David Vandyke, and Steve~J. Young. 2016.
\newblock \href {http://aclweb.org/anthology/N/N16/N16-1015.pdf} {Multi-domain
  neural network language generation for spoken dialogue systems}.
\newblock In \emph{{NAACL} {HLT} 2016, The 2016 Conference of the North
  American Chapter of the Association for Computational Linguistics: Human
  Language Technologies}, pages 120--129, San Diego, CA, USA.

\bibitem[{Wen et~al.(2015{\natexlab{b}})Wen, Gasic, Mrksic, Su, Vandyke, and
  Young}]{SemanticLSTMNLG_Wen15}
Tsung{-}Hsien Wen, Milica Gasic, Nikola Mrksic, Pei{-}hao Su, David Vandyke,
  and Steve~J. Young. 2015{\natexlab{b}}.
\newblock {Semantically Conditioned LSTM-based Natural Language Generation for
  Spoken Dialogue Systems}.
\newblock In \emph{Proceedings of the 2015 Conference on Empirical Methods in
  Natural Language Processing, {EMNLP} 2015}, pages 1711--1721, Lisbon,
  Portugal.

\bibitem[{Wiseman et~al.(2017)Wiseman, Shieber, and
  Rush}]{ChallengesInData2DocGeneration_Wiseman17}
Sam Wiseman, Stuart Shieber, and Alexander Rush. 2017.
\newblock \href {https://aclanthology.info/papers/D17-1239/d17-1239}
  {Challenges in data-to-document generation}.
\newblock In \emph{Proceedings of the 2017 Conference on Empirical Methods in
  Natural Language Processing, {EMNLP} 2017}, pages 2253--2263, Copenhagen,
  Denmark.

\bibitem[{Wu et~al.(2016)Wu, Schuster, Chen, Le, Norouzi, Macherey, Krikun,
  Cao, Gao, Macherey, Klingner, Shah, Johnson, Liu, Łukasz Kaiser, Gouws,
  Kato, Kudo, Kazawa, Stevens, Kurian, Patil, Wang, Young, Smith, Riesa,
  Rudnick, Vinyals, Corrado, Hughes, and Dean}]{NMTGoogle_Wu16}
Yonghui Wu, Mike Schuster, Zhifeng Chen, Quoc~V. Le, Mohammad Norouzi, Wolfgang
  Macherey, Maxim Krikun, Yuan Cao, Qin Gao, Klaus Macherey, Jeff Klingner,
  Apurva Shah, Melvin Johnson, Xiaobing Liu, Łukasz Kaiser, Stephan Gouws,
  Yoshikiyo Kato, Taku Kudo, Hideto Kazawa, Keith Stevens, George Kurian,
  Nishant Patil, Wei Wang, Cliff Young, Jason Smith, Jason Riesa, Alex Rudnick,
  Oriol Vinyals, Greg Corrado, Macduff Hughes, and Jeffrey Dean. 2016.
\newblock \href {http://arxiv.org/abs/1609.08144} {Google's neural machine
  translation system: Bridging the gap between human and machine translation}.
\newblock \emph{CoRR}, abs/1609.08144.

\bibitem[{Zhou et~al.(2017)Zhou, Lapata, Wei, Dong, Huang, and
  Xu}]{LearningToGenerateProductReviews_Zhou17}
Ming Zhou, Mirella Lapata, Furu Wei, Li~Dong, Shaohan Huang, and Ke~Xu. 2017.
\newblock Learning to generate product reviews from attributes.
\newblock In \emph{Proceedings of the 15th Conference of the European Chapter
  of the Association for Computational Linguistics, {EACL} 2017, Volume 1: Long
  Papers}, pages 623--632, Valencia, Spain.

\end{thebibliography}
\bibliographystyle{acl_natbib}

\appendix
\section{Hyperparameters for Models Trained on Synthetic Training Data}
For the model trained on template-generated data, we tune the hyperparameters to achieve 100\%~accuracy for their best hypotheses on template-generated references on the development set.
All models have a single-layer LSTM with 64~hidden units in the encoder and decoder.
We half the learning rate starting from the eighth training epoch or if the perplexity of the validation set does not improve. The gradient norm is capped at two.
The decoder uses the general attention mechanism.
For decoding, we set the beam size to~30.
Table~\ref{tab:model_hyperparams} shows hyperparameters which differ for the models.

\begin{table}[h!tb]
	\centering
	  \begin{tabularx}{\linewidth}{p{3cm}|XXX}
      hyperparameter & T 1 & T 2 & T 1+2 \\
      \midrule
      encoder & \multicolumn{2}{c}{unidirectional} & bidir.\\
      embedding size & 28 & 28 & 30\\
      optimizer & Adam & SGD & SGD\\
      init. learning rate & 0.001 & 1.0 & 1.0\\
      batch size & 4 & 4 & 16\\
          dropout & 0.4 & 0.5 & 0.3\\
	epochs & 25 & 13 & 15\\
      \end{tabularx}
      \caption{Hyperparameters for the models trained on synthetic training data generated from Template~1~(T~1), Template~2~(T~2) and both (T~1+2).}
      \label{tab:model_hyperparams}
\end{table}

\end{document}